\documentclass[10pt,emptycopyrightspace]{ewsn-proc}

\usepackage{graphicx}
\usepackage{balance}
\usepackage{comment}

\usepackage[utf8]{inputenc}
\usepackage{fourier} 
\usepackage{array}
\usepackage{makecell}

\numberofauthors{1}
\author{
\alignauthor Mohammad Arif Ul Alam\\
  \affaddr{University of Massachusetts Lowell\\
  Lowell, MA, USA\\
  mohammadariful\_alam@uml.edu\\
  University of Massachusetts Chan Medical School\\
  Worcester, MA, USA\\
  mohammadariful.alam1@umassmed.edu}
}

\title{Internet of Things Fault Detection and Classification via Multitask Learning}

\begin{document}

\maketitle

\begin{abstract}
This paper presents a comprehensive investigation into developing a fault detection and classification system for real-world IIoT applications. The study addresses challenges in data collection, annotation, algorithm development, and deployment. Using a real-world IIoT system, three phases of data collection simulate 11 predefined fault categories. We propose SMTCNN for fault detection and category classification in IIoT, evaluating its performance on real-world data. SMTCNN achieves superior specificity (3.5\%) and shows significant improvements in precision, recall, and F1 measures compared to existing techniques.
\end{abstract}

\section{Introduction}
The Industrial Internet of Things (IIoT) refers to the network of interconnected devices, sensors, and systems in industrial settings that enable the collection, exchange, and analysis of data to enhance operational efficiency and productivity \cite{i1,i2}. It involves integrating physical machinery and equipment with digital technologies, enabling real-time monitoring, control, and automation of industrial processes \cite{i3}. However, within the complex IIoT ecosystem, various faults and anomalies can occur, including equipment malfunctions, communication failures, cybersecurity breaches, and data inaccuracies. Detecting and addressing these faults is crucial for maintaining uninterrupted operations, ensuring worker safety, minimizing downtime, and optimizing resource utilization \cite{i4}.

Detecting and predicting anomalies in industrial environments is crucial for economic and security purposes. However, the rarity of these events poses challenges when applying existing algorithms, leading to false alarms or misdetections \cite{i6}. Previous studies conducted by Nardelli et al. \cite{i7}, Leitão et al. \cite{i8}, and Oks et al. \cite{i9} have explored the modeling of IIoT networks and cyber-physical systems in industrial settings. Notably, promising solutions have been presented in reference \cite{19}. Fault detection in industrial environments has always been challenging, with difficulties arising from device interoperability and limitations in data collection. The most promising state-of-the-art method utilizes Generative Adversarial Networks (GANs) for the detection and classification of system failures using a dataset containing missing values \cite{state_of_art}. However, the focus of our paper is to develop a simple yet powerful unified method for the real-time detection and classification of IIoT faults using data collected from a real-world system.

To date, the existing literature on IIoT fault detection and classification has predominantly focused on either anomaly detection or fault classification for industrial processes. In this paper, we propose a novel framework called Sequential Multitask Cascaded Neural Network (SMTCNN) that integrates various approaches for anomaly detection and classification. The concept of Multitask Cascaded Neural Network (MTCNN) has gained significant popularity in face recognition research \cite{i5}, where it involves three cascaded tasks: detecting the face area, detecting facial features such as eyes, nose, and mouth, and finally speeding up the face detection process. We adapt the MTCNN framework for sequential data, resulting in SMTCNN, which consists of three tasks: change-point detection, anomaly detection, and fault classification. The framework is designed to effectively handle small fault event datasets and large normal datasets using a cascaded learning scheme. Our work makes the following key contributions:
\begin{itemize}
    \item We developed a novel Sequential Multitask Cascaded Neural Network (SMTCNN) architecture, which incorporates three sequential tasks: change-point detection, refinement of change-point detection for anomaly/fault event identification, and classification of the detected faulty events using simple neural networks.
    \item We devised a comprehensive data collection methodology encompassing normal, abnormal, and real-world scenarios to gather IIoT data from a deployed system. Subsequently, we collected the data and conducted an evaluation of our proposed SMTCNN model. Our experimental results demonstrate the superior performance of our approach compared to existing state-of-the-art methods, particularly in terms of specificity (3.5\%).
\end{itemize}

\begin{figure*}[!htb]
\begin{center}
 \includegraphics[width=0.7\linewidth]{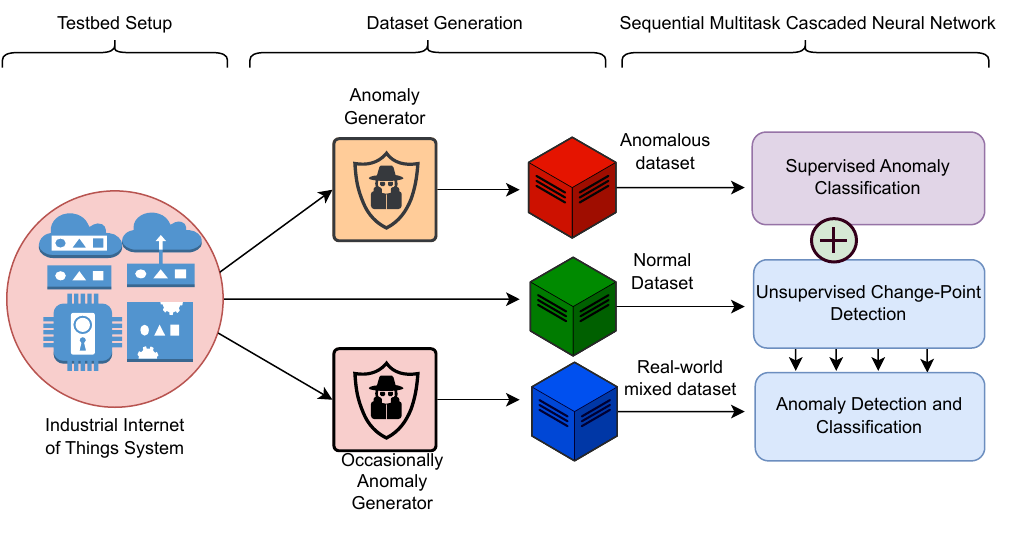}
 \caption{The schematic diagram of proposed framework}
 \label{fig:overview}
\end{center}
\end{figure*}

\section{Testbed Setting}
The four IIoT devices consisted of an industrial application module and an IIoT device monitoring agent module. Two devices monitored the oven temperature, while the other two monitored smoke and humidity levels, all at a frequency of once per minute. Data from the devices were transmitted to the network gateway via WirelessHART for alarm functionality. The IIoT device monitoring agent captured hardware and firmware metrics, which were relayed to the gateway. Each device had sensors, a radio, microcontroller unit (MCU), and power supply, with the radio using WirelessHART for communication with the MCU. The gateway managed network operations, security tasks, and connected the IWSN to an IP network.
\section{Datasets Generation}
The dataset utilized in this research was created by executing the firmware of the IIoT devices for an extended duration and capturing the corresponding metrics for each function in a dedicated database. To evaluate the system's behavior in the presence of faults, deliberate firmware and hardware anomalies were introduced into the WirelessHART testbed at predetermined intervals. These anomalies were carefully designed to emulate real-world situations and assess the resilience and efficacy of the fault detection and classification algorithms. The injected faults encompassed 11 distinct fault classes that can be classified into four distinct types (Table \ref{tab:types}).

\begin{table}[h]
\addtolength{\tabcolsep}{-4pt}
  \caption{Different types of IIoT faults with description}
  \label{tab:types}
 \centering
 \begin{tabular}{|p{3cm}|p{.8cm}|p{4cm}|}
    \hline
     {\bf Types} & {\bf Class} & {\bf Description}\\
    \hline
    Under voltage faults & \makecell{ 1\\2\\3\\4\\5\\6} & \makecell{ with 3.3-3.0v\\with 3.3-2.8v\\with 3.3-2.6v\\with 3.3-2.4v\\with 3.3-2.3v\\with 3.3-2.2v} \\
    \hline
    Sensor stuck-faults by stimulating SPI& \makecell{ 7\\8} & \makecell{ with VCC temperature\\with Clock}\\
    \hline
    High temperature faults in microcontroller unit & \makecell{ 9 \\ 10}  & \makecell{ under high voltages \\ under low voltages}  \\
    \hline
    Buffer overflow faults & \makecell{11} & A buffer overflow vulnerability used for running arbitrary code\\
    \hline
  \end{tabular}
\end{table}

By employing the aforementioned fault generation technique, three distinct datasets were generated, each comprising time logs, energy consumption, CPU usage, and time fields (representing the duration of each instance in seconds) for every data point.

\begin{itemize}
    \item {\bf Anomaly Only Dataset}: To collect this small dataset, a series of continuous anomalous event were deliberately induced in the IIoT system. The dataset specifically targets 11 distinct categories of anomalies, aiming to provide comprehensive coverage of potential abnormal behaviors within the system.
    \item {\bf Normal Only Dataset}: This dataset was generated in a normal operational environment of IIoT, where no known faults were present.
    \item {\bf Real-time Mixed Dataset}: This dataset comprises a fully operational real-time deployed IIoT system, where randomly chosen faults from the aforementioned 11 categories were sporadically introduced (1\% of the time during the day) to simulate a realistic real-world scenario.
\end{itemize}

\section{IIoT Faults Detection and Classification}
\subsection{Unsupervised Change-Point Detection using Simple LSTM Autoencoder}
We have developed an unsupervised change-point detection algorithm utilizing an autoencoder that is based on the Long Short-Term Memory (LSTM) neural network. Our approach was inspired by the work of Elhalwagy et al. \cite{capsule}, who proposed a hybridization of unsupervised LSTM and Capsule network for online change-point detection in time-series data. The architecture of our autoencoder model consists of two LSTM-based encoders, one for each channel of the input data. The encoded representations from both channels are concatenated and passed to the decoder. The decoder's role is to reconstruct the input data, and during training, the model is optimized using the mean squared error (MSE) loss function and the Adam optimizer. To mitigate overfitting, we employed early stopping during training and restored the weights associated with the best performance. Subsequently, the trained autoencoder is utilized to reconstruct the time series data. Change-points are identified by establishing a threshold, denoted as $\tau$, which is determined based on the mean and standard deviation of the reconstruction errors. Any reconstruction error exceeding the threshold is considered a change-point. The algorithm effectively detects change-points, signifying significant variations or anomalies in the sequential data that may indicate the occurrence of faulty events in IIoT.
\subsection{Supervised IIoT Faults Classification}
Due to the limited size of the collected IIoT faults dataset, we employed state-of-the-art algorithms, excluding neural networks, for fault classification. Specifically, we utilized Random Forest (RF), Support Vector Machines (SVM), Naive Bayes (NB), Logistic Regression (LG), Decision Tree (DT), and Stochastic Gradient Descent (SGD) algorithms. To effectively apply these algorithms, we employed suitable segmentation and windowing techniques.
\subsection{Sequential Multitask Cascaded Neural Networks (SMTCNN)}
\begin{figure}[!htb]
\begin{center}
 \includegraphics[width=\linewidth]{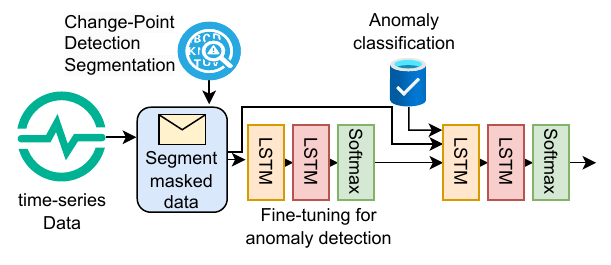}
 \caption{Proposed Sequential Multitask Cascaded Neural Network Architecture for IIoT faults detection and classification}
 \label{fig:SMTCNN}
\end{center}
\end{figure}

Fig. \ref{fig:SMTCNN} shows our overall SMTCNN model architecture for IIoT Fault detection and classification.

{\bf Problem formulation}: Let $X$ be the input feature matrix with three columns. The binary label set $Y_{{anomaly}}$ indicates the presence or absence of an anomaly, while the 12-class set $Y_{{class}}$ represents different fault types, with label 12 indicating no fault and labels 1 to 11 corresponding to 11 specific fault types. We define three tasks: $T_1$, $T_2$, and $T_3$, each with their respective output variables $O_{{t1}}$, $O_{{t2}}$, and $O_{{t3}}$.

{\bf Task 1: Anomalous segment identification}: This approach takes inspiration from the original Multitask Cascaded Neural Network (MTCNN) algorithm \cite{i5}, originally designed for face recognition. Specifically, we focus on Task-1 of the MTCNN algorithm, which is responsible for detecting bounding boxes around faces. In the context of our change-point detection algorithm for real-time sequential data, we adapt this concept to identify the start and end points of change-points, similar to bounding boxes in the temporal domain. These identified change-points act as markers to segment the data into regions of interest that may contain anomalous events requiring further analysis. The underlying hypothesis behind this segmentation approach is that these specific regions contain valuable information related to anomalous events and merit closer examination.

{\bf Task 2: Fine-tuning anomaly detection}: This step considers the potential segments predicted by Task 1 and aims to predict anomalies. For this purpose, we employ a simple two-layer LSTM network followed by a fully connected layer with two outputs. This network takes the original features as input.

{\bf Task 3: Anomaly detection and classification}: This task involves the final refinement of anomaly detection and classification using the original feature set $X$, the output of Task 1 ($O_{t1}$) which represents the segmented mask on the time-series data indicating detected change-points, and the output of Task 2 ($O_{t2}$) which represents the fine-tuned detected anomalies. The core network for Task 3 is also a simple two-layer LSTM network with a fully connected neural network that uses softmax activation and has an output dimension of 12 (representing 11 different fault categories and one label for the existence of a fault). Thus, the final loss function is defined as follows:

\begin{equation}
Loss_{SMTCNN} = -\frac{1}{N} \sum_{i=1}^{N} \sum_{t=1}^{T} \sum_{c=1}^{C} y_{i,t,c} \log(p_{i,t,c})
\end{equation}

In the above equation, $N$ is the number of samples, $T$ is the number of time steps in the sequential data, $C$ is the number of classes, $y_{i,t,c}$ is the one-hot encoded ground truth label for sample $i$, time step $t$, and class $c$, and $p_{i,t,c}$ is the predicted probability of sample $i$ at time step $t$ belonging to class $c$.
\section{Experimental Evaluation}
The proposed SMTCNN framework was implemented using Python, TensorFlow, and scikit-learn. For our study, we collected three datasets: 8,432 data points of anomalous data, 740,448 data points of normal data, and 718,444 data points of a mixed dataset representing real-time scenarios. The unsupervised change-point detection algorithm was trained using the normal data only, while the supervised IIoT faults classification algorithm was trained using the anomalous data only. Finally, the SMTCNN algorithm was trained and evaluated using the mixed dataset, encompassing both normal and anomalous data.
\subsection{Baseline}
We have implemented few state-of-art algorithms as well as few versions of our proposed framework by removing few modules for prooving their importance in the pipeline.
\begin{itemize}
    \item {\bf B1 (GAN based Method)}: This method \cite{state-of-art} proposed a GAN-based anomaly detection and classification algorithm to detect and classify faults in IIoT.
    \item  {\bf B2 (SMTCNN without change-point-detection segmentation)}: In this framework, we removed the change-point detection based segmentation masking, but kept all other modules.
    \item  {\bf B2 (SMTCNN without supervised faults classification)}: In this framework, we removed supervised faults classification module in the SMTCNN entwork, but kept every other modules.
\end{itemize}
\subsection{Results}
We employed the traditional 10-fold cross-validation technique to evaluate the supervised segmented IIoT fault classification. However, as traditional 10-fold cross-validation is not suitable for sequential data, we adapted our approach. To train and assess the performance of our proposed sequential algorithm, SMTCNN, we divided the entire sequential data into two halves. Subsequently, we randomly selected a sequence of data from the first half for training purposes and another sequence of data from the second half for testing. The lengths of the training and testing sequences were also randomly chosen within a range of 50\% to 80\% of the available data samples for each fold. This process was repeated 10 times to generate 10 different sets of training and testing sequences. We evaluated our proposed algorithm's performance using metrics such as balanced accuracy, precision, recall, specificity, and F1 measure. Additionally, we calculated the standard deviation of these metrics to assess overfitting. The performance details of the baselines and our overall framework are provided in Table \ref{tab:overall_accuracy}, clearly demonstrating the superior performance of our framework compared to the baseline algorithms.

We observed that our framework exhibits marginal improvements in terms of accuracy, precision, recall, and F-1 measure compared to the baseline algorithms. However, the improvement in specificity is particularly noteworthy, with a significant increase of 3.5\% over the baseline (B1). This enhancement represents a substantial advancement in the field of IIoT fault detection research. A closer examination of Table \ref{tab:overall_accuracy} reveals that the performance of SMTCNN without change-point detection segmentation (B1) and without supervised fault classification pretraining (B2) experiences a significant drop in performance.

\begin{table}[h]
\addtolength{\tabcolsep}{-4pt}
  \caption{Performance comparisons of our proposed method with baseline algorithms}
  \label{tab:overall_accuracy}
 \centering
 \begin{tabular}{|p{0.5cm}|p{1.4cm}|p{1.4cm}|p{1.4cm}|p{1.5cm}|p{1.3cm}|}
    \hline
     &  Accuracy & Precision & Recall & Specificity & F1-measure\\ 
    \hline
    B1 & 96.9$\pm$.02 & 96.6$\pm$.02 & 96.9$\pm$.02 & 96.1$\pm$.01 & 97.4$\pm$.02\\

    B2 & 94.5$\pm$.05 & 96.3$\pm$.03 & 94.5$\pm$.05 & 96.8$\pm$.02 & 96.5$\pm$.01\\

    B3 & 95.1$\pm$.04 & 94.6$\pm$.05 & 95.1$\pm$.04 & 97.1$\pm$.03 & 97.1$\pm$.01\\

    Our & 97.9$\pm$.01 & 96.8$\pm$.02 & 97.9$\pm$.01 & 98.6$\pm$.01 & 98.1$\pm$.01\\
    \hline
  \end{tabular}
\end{table}

% \begin{figure}[!htb]
% \begin{center}
%  \includegraphics[width=\linewidth]{faults_results.pdf}
%  \caption{Performance comparisons of each baseline algorithm with ours on each fault classification metrics}
%  \label{fig:faults_results}
% \end{center}
% \end{figure}

\subsection{Conclusions and Future Works}
This paper represents an initial step towards the collection, simulation, and prediction of faults in real-world deployed IIoT systems. It is a part of our broader vision to develop a secure, fault-tolerant, and reliable smart and connected industrial infrastructure. The primary objective of this study is to release our preliminary collected datasets for the advancement of IIoT fault generation and the integration of multitask learning in fault detection. However, it is essential to acknowledge that automatic fault detection encompasses a wide range of use cases that were not specifically addressed in this research.

In the context of IIoT, it is crucial to consider the diversity among different facilities, which may necessitate specific modifications to the fault detection algorithm by incorporating scalable and adaptable machine learning techniques. Moreover, in real-world scenarios, IIoT systems often encounter a significant number of missing values resulting from internet connectivity issues or power interruptions, which were not explicitly tackled in this study.

\end{document}